\def\year{2021}\relax
\documentclass[letterpaper]{article} 
\usepackage{aaai21}  
\usepackage{times}  
\usepackage{helvet} 
\usepackage{courier}  
\usepackage[hyphens]{url}  
\usepackage{graphicx} 
\usepackage{natbib}  
\usepackage{caption} 
\title{Self-supervised Pre-training and Contrastive Representation Learning\\for Multiple-choice Video QA}

\author{
    Seonhoon Kim\textsuperscript{\rm 1,2},
    Seohyeong Jeong\textsuperscript{\rm 1},
    Eunbyul Kim\textsuperscript{\rm 2},
    Inho Kang\textsuperscript{\rm 2},
    Nojun Kwak\textsuperscript{\rm 1,$\dagger$} \\
}
\affiliations{
    \textsuperscript{\rm 1} Seoul National University,
    \textsuperscript{\rm 2} Naver Search\\
    \{seo\_hyeong, nojunk\}@snu.ac.kr, 
    \{seonhoon.kim, silverstar.kim, once.ihkang\}@navercorp.com \\
}

\usepackage[switch]{lineno}
\usepackage{multirow}
\usepackage{amsmath}
\usepackage{amssymb}
\DeclareMathOperator*{\argmax}{argmax}
\usepackage{subfig}
\usepackage{color}

\newcommand{\nj}[1]{{\color{black}{#1}}}
\newcommand{\sh}[1]{{\color{black}{#1}}}
\newcommand{\ksh}[1]{{\color{black}{#1}}}

\begin{document}
\maketitle

\begin{abstract}
Video Question Answering (Video QA) requires fine-grained understanding of both video and language modalities to answer the given questions. In this paper, we propose novel training schemes for multiple-choice video question answering with a self-supervised pre-training stage and a supervised contrastive learning in the main stage as an auxiliary learning. In the self-supervised pre-training stage, we transform the original problem format of predicting the correct answer into the one that predicts the relevant question to provide a model with broader contextual inputs without any further dataset or annotation. For contrastive learning in the main stage, we add a masking noise to the input corresponding to the ground-truth answer, and consider the original input of the ground-truth answer as a positive sample, while treating the rest as negative samples. By mapping the positive sample closer to the masked input, we show that the model performance is improved. We further employ locally aligned attention to focus more effectively on the video frames that are particularly relevant to the given corresponding subtitle sentences. We evaluate our proposed model on highly competitive benchmark datasets related to multiple-choice video QA: TVQA, TVQA+, and DramaQA. Experimental results show that our model achieves state-of-the-art performance on all datasets. We also validate our approaches through further analyses.

\end{abstract}

\section{Introduction}

\let\thefootnote\relax\footnote{
$\dagger$ Corresponding author.}

\begin{figure}[t]
  \includegraphics[width=\linewidth]{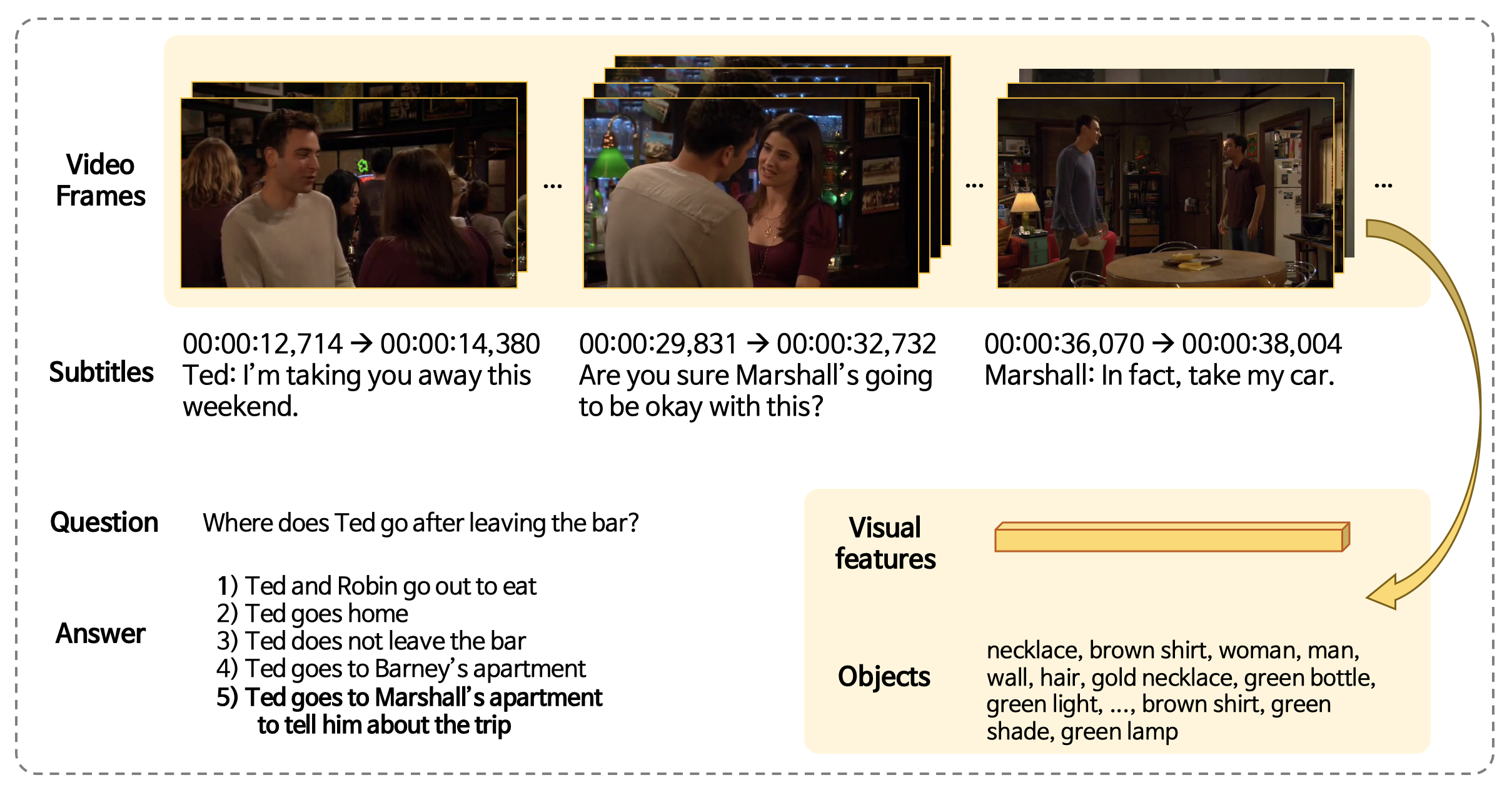}
  \caption{Multiple-choice Video QA example of TVQA dataset, composed of a 60-90 second long video clip, question, and the answer options. A video clip consists of video frames and subtitles, and each subtitle is connected to several frames. In our setting, we additionally extract object information and visual features from the video frames using Faster R-CNN and ResNet-101 as in the bottom right yellow box. We use question, answer, subtitles, and objects as our text input and visual features as our visual input.} 
\label{fig:example}
\end{figure}

Recent years have witnessed significant improvements in vision and language communities, which have consequently led to substantial attention in vision-language multi-modality tasks such as visual grounding \cite{Plummer_2015_ICCV}, image captioning \cite{chen2015microsoft}, and visual question answering \cite{antol2015vqa, goyal2017making}. Furthermore, as video becomes ubiquitous, as a daily source of information and communication, video-language tasks such as video captioning \cite{zhou2018end}, video moment retrieval \cite{liu2018attentive}, and video question answering (video QA) \cite{lei2018tvqa, lei2019tvqa+} are emerging as important topics. Among these topics, video QA is especially challenging, as it requires fine-grained understanding of both video and language. \sh{Figure \ref{fig:example} shows an example of multiple-choice video QA from the TVQA dataset. The multiple-choice video QA task requires the model to select the correct answer given a question, corresponding video frames, and subtitles.}

To address video QA, several works have utilized early stage fusion method \cite{kim2017deepstory, na2017read} to merge two different modalities, while other recent works \cite{lei2018tvqa,lei2019tvqa+} have employed late stage fusion method, which extracts representation from language and vision independently during the early stage of framework, and combines them in QA-aware manners. To obtain further fine-grained information from videos, \nj{\citet{kim2020densecaption}} has generated captions using a dense caption model \cite{yang2016dense} to translate vision modality to that of language. Furthermore, \citet{geng2020character} has reckoned that explicitly replacing the predicted regions corresponding to a person from object detection with the name of protagonists helps the model to answer the questions. 

Unlike the previously mentioned methods, which focus on extracting QA-aware visual information, we shift our focus to the training procedure that could possibly take the most advantage out of the given data. On top of utilizing
large-scale pre-trained language model and fine-grained object detection results on videos, motivated by recent progress in contrastive learning \cite{khosla2020supervised,chen2020simple} and unsupervised pre-training in natural language processing \cite{geng2020character}, we propose a training scheme for multiple-choice video QA that integrates these two perspectives to increase performance gain.

Our framework consists of two consecutive stages of training\nj{:} first is the self-supervised pre-training stage and second is the training stage with a supervised contrastive learning loss in
an auxiliary loss setting. During the self-supervised pre-training, \ksh{instead of predicting the correct answer,} \nj{our} model is expected to predict the relevant question given contexts such as video clips and subtitles to learn a better weight initialization. 
\ksh{This procedure does not require any additional data or human annotation.}
\ksh{For the fine-tuning stage, in addition to the main QA loss, we propose a contrastive loss that can be applied for the multiple-choice video QA tasks and we also make use of an optional span loss.}
Taking ground truth answer as a positive sample and the rest as negative samples, \nj{the} contrastive loss confines the positive sample to be mapped in the neighborhood of an anchor, \nj{a} perturbed ground truth answer, and the negative samples to be away from the anchor.
\ksh{We further show the effectiveness of contrastive loss by investigating how the distance between the positive sample and negative samples changes as the training continues.}

In addition, we present a locally aligned attention mechanism to selectively extract video representations corresponding to the given subtitles. Previous works \cite{lei2018tvqa,lei2019tvqa+,kim2020densecaption} have utilized attention mechanisms on video sequences and subtitles \sh{with either} question and answer pairs respectively or with the subtitles in a globally aligned manner. In contrast, we hypothesize that it is desirable to apply a direct attention mechanism that computes attention score between two modalities in locally aligned fashion. Performing attention in locally aligned fashion is beneficial to the model's performance, since it prevents the model from reasoning with unnecessary information.

We evaluate the proposed approach on large-scale TV shows-based question answering datasets; TVQA, TVQA+, and DramaQA. Each video clip is paired with corresponding subtitles and natural language multiple-choice questions.
Empirically, our model takes advantage of the supervised contrastive loss in the main stage and gives further improvements when self-supervised pre-training is preceded.
Moreover, our model demonstrates significant performance increase on the test server, outperforming the state-of-the-art scores. Our contributions are as follow:

\begin{itemize}
\item We improve the performance of the model by adding a novel self-supervised pre-training stage.
\item We introduce additional supervised contrastive loss in an auxiliary setting during the main stage of training.
\item We propose a locally aligned attention mechanism to selectively focus on corresponding video sequences of given subtitles.
\item We show that our framework achieves state-of-the-art performance on TVQA, TVQA+, and DramaQA.
\end{itemize}

\begin{figure*}[t]
  \centering
  \includegraphics[width=0.83\linewidth]{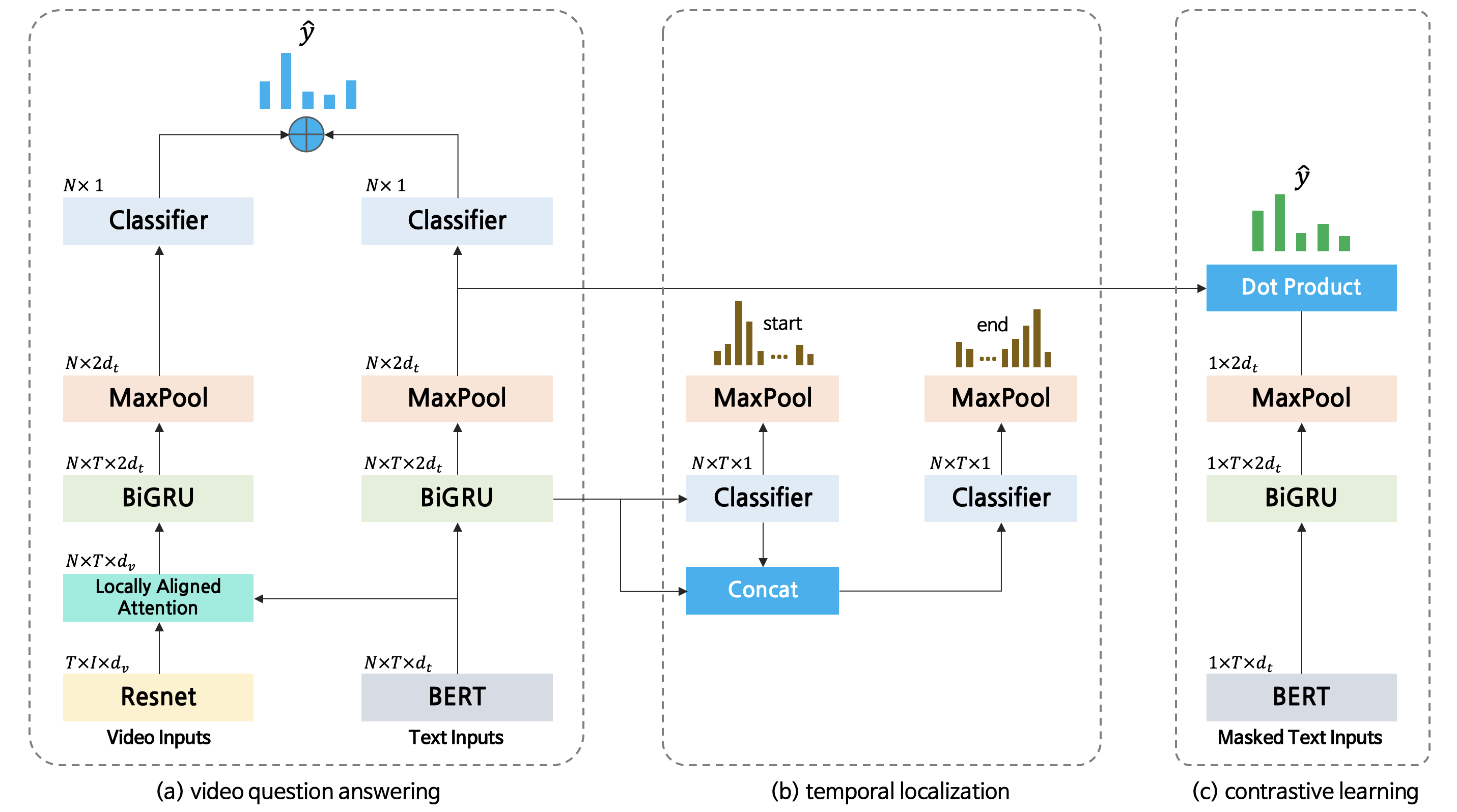}
  \caption{Overall architecture of our model: (a) For a video QA part, we use ResNet and BERT to extract video and text representations. A locally aligned attention mechanism is introduced to match each subtitle sentence with the corresponding images.
  Then, we use RNNs to learn sequential information of subtitle sentences. We predict the final answer distribution on both modalities. At inference time, we use this video QA part only. (b) Temporal localization, one of our auxiliary tasks, is used to predict the necessary part to answer the question. 
  (c) We introduce the contrastive loss, which is another component of our auxiliary tasks, to enhance the model's performance. We utilize the identical BERT and RNN, used in a video QA part with the masked text input of the ground-truth and predict the answer distribution by contrasting positive pair against negative pairs.}  
  \label{fig:arch}
\end{figure*}

\section{Related Work}
\subsection{Visual/Video Question Answering}

Visual and video question answering requires the fine-grained interplay of vision and language to understand multi-modal contents. 
In the last few years, most of the pioneering works used a single image as a visual content with a joint image-question embedding and a spatial attention to predict \sh{the correct} answer \cite{antol2015vqa,xu2016ask,Yang_2016_CVPR,fukui2016multimodal}. More recently, beyond question answering on a single image, as video has become an important source of information, video QA has emerged as a key topic in the vision-language community \cite{lei2018tvqa, lei2019tvqa+, kim2019progressive, thaominh2019, kim2020densecaption, kim2020modality, geng2020character}. Contrast to previous works done in video QA, in this work, we focus not only on learning the multi-modal representations, but also on the training procedure that could take additional advantages of the given dataset.

\subsection{Self-supervised Learning}

Modern techniques of self-supervised learning \sh{are} pre-trained on large-scale external and unlabeled datasets \cite{devlin2018bert,liu2019roberta,raffel2019exploring}.
Several studies \cite{lei2019tvqa+,Yang_2020_WACV,kim2020densecaption,kim2020modality} have taken advantages of these self-supervised pre-trained models and combined with their video QA models to learn representations of the text data such as question\sh{s}, answers, subtitles, and extracted visual concepts.
Likewise, we utilize the pre-trained \sh{language model to embed textual information to solve video QA tasks.} 
Besides, we \nj{propose} a self-supervised learning approach for multiple-choice video QA \sh{of predicting a relevant question given contexts}, which does not require any additional data or further annotations.

\subsection{Contrastive Representation Learning}

Contrastive representation learning \cite{hadsell2006dimensionality} has been explored in numerous literature as a method of extracting powerful feature representation. The main goal of the learning is to, as the name suggests, contrast the semantically nearby points against dissimilar points, in the embedding space. Many tasks \cite{dosovitskiy2014discriminative, he2020momentum, chen2020simple} have been proposed to incorporate various forms of contrastive loss into a self-supervised learning algorithm. Meanwhile, some approaches \cite{khosla2020supervised, tian2020makes} have focused on leveraging labeled data into contrastive representation learning. In this work, we exploit supervised contrastive loss in an auxiliary setting on top of the \ksh{main} task to learn the better representation.

\section{Methods}
In this section, we describe our architecture for multiple-choice video QA \sh{with an additional auxiliary learning task using a contrastive loss.}
\sh{Furthermore}, a self-supervised pre-training approach can be applied before the main training stage. For our problem setting, the inputs are composed of the following: (1) question $q$, (2) answer options \nj{$\Omega_a = \{a_n | n=1,...,N\}$}, (3) subtitle sentences $\{S_t |t=1,...,T\}$ as a text context, and (4) video frames $\{V_t^{i} |t=1,...,T, i=1,...,I \}$ as a visual context where $a_n$ is the $n^{th}$ answer option, $S_t$ is the $t^{th}$ subtitle sentence, and $V_t^{i}$ is the $i^{th}$  image frame in the $t^{th}$ video segment connected to the $t^{th}$ subtitle sentence. Our goal is to predict the correct answer given a question and text/visual contexts.

\begin{equation} 
\hat{a} = \argmax_{a\in \Omega_a} p(a|q,S,V;\theta) 
\label{eq:problem}
\end{equation}

\subsection{Input Representation}

\subsubsection{Visual representation}
We first separate each video into $T$ segments using the \sh{provided subtitle timestamp in the dataset,} and further separate each segment into $I$ image frames. 
Then, \sh{as shown in Figure \ref{fig:example}, for the visual representation,} we use ResNet-101 \cite{he2016deep} trained on ImageNet \cite{deng2009imagenet} to extract global image features $v_t^{i}\in \mathbb{R}^{2048}$ as the $i^{th}$ image feature in the $t^{th}$ video segment. 
In addition, using Faster R-CNN \cite{ren2015faster} trained on Visual Genome \cite{krishna2017dense}, we extract objects $o_t^{i_j}$ as $j^{th}$ object in the $i^{th}$ image frame, which can be used as one of the text inputs described in the next subsection.

\subsubsection{Text representation}
We use four types of text inputs: a question, answer options, subtitle sentences, and objects. For the objects $o_t$, extracted from each image frame in the $t^{th}$ video segment, \sh{we use the following as the objects input: }

\begin{equation} 
o_t = [o_t^{1_1};...;o_t^{1_{J_1}};...;o_t^{I_1};...;o_t^{I_{J_I}}]
\end{equation}
where [\textperiodcentered;\textperiodcentered] is the concatenation operator. To encode entire textual inputs, we use BERT \cite{devlin2018bert} which achieves state-of-the-art performance on a wide range of NLP tasks. 
While only one or two types of text inputs are used with $\mathtt{[SEP]}$ tokens in the standard practices of BERT, \nj{since we use four different types of text inputs, we separate} them with $\mathtt{[SEP]}$ tokens as follows:

\begin{quote}
\centering 
$\mathtt{[CLS]}$ $q$ $\mathtt{[SEP]}$ $a_n$ $\mathtt{[SEP]}$ $S_t$ $\mathtt{[SEP]}$ $o_t$ $\mathtt{[SEP]}$.
\end{quote}
To properly distinguish multiple text inputs (four in our case) in the model, we modify the token type embedding method \sh{to explicitly accommodate different token type embeddings as types of text inputs vary.}
For the first input, we keep the token type embedding of $0$ as it is. For the second and third inputs, we use the output of the token type embedding of $1$ but multiplied by $\frac{1}{3}$ and $\frac{2}{3}$, respectively, to distinguish them. Lastly, we keep the token type embedding of $1$ for the fourth text input.

\subsection{Model}
Our model consists of two stages: 
1) self-supervised pre-training stage with the transformed problem format, and
2) video QA as a main stage.
For the video QA stage, in addition to predicting the answer as our main task, we \sh{make use of timestamp annotation of localized span needed to answer the question given in the dataset and add temporal localization learning as an auxiliary task. And the supervised contrastive learning is combined as an auxiliary task in the main training \nj{stage} as well.} 
In the next subsections, we first describe our video QA architecture (Fig. \ref{fig:arch}), and then, we describe how we utilize the self-supervised pre-training as our prerequisite learning.

\subsubsection{Video Question Answering}

We use visual and text inputs for our video QA network as shown in Fig. \ref{fig:arch}(a).
For the visual representation $H_v\in \mathbb{R}^{T \times I \times d_v}$, we extract $d_v=2048$ features of the last block of ResNet-101 which was used in \citet{lei2018tvqa}. We set the number of images $I$ as 4, extracted from the video segment connected to each subtitle sentence. \ksh{In our implementation, we repeated $H_v$ $N$ times to match the dimension with the text representation.}
For the text representation $H_t\in \mathbb{R}^{N \times  T \times d_t}$, we extract $d_t=768$ features of the hidden state of the $\mathtt{[CLS]}$ token \sh{from} the last layer of 12-layer BERT-base model.

To extract attentive information between \nj{a text context from a subtitle and a visual context from the corresponding video frames}, we calculate the locally aligned attention to focus on \sh{particularly} relevant images regarding each subtitle sentence. \sh{This prevents} the model from reasoning with unnecessary information.
Our locally aligned attention mechanism, used only in the image side, is calculated between image frames and \nj{the subtitle sentence} that share the timestamp with the image frames.
\begin{equation}
\begin{split}
H_v^{Att} = \sum_{i=1}^{I}  \alpha_i H_{v_i}^T\mathbf{M}, \  
\alpha_i = \frac{e^{g_i}}{\sum_{k=1}^{I} e^{g_k}}, \
g_i = H_{v_i}^T\mathbf{M}H_t.
\end{split}
\label{eq:attention}
\end{equation}

Here, $I$ is the number of image frames from the video segment \sh{matching to} each subtitle sentence by the \sh{timestamp information given in the dataset}, and $\mathbf{M}$ is the \ksh{projection matrix} that converts the text representation into the visual representation space.

To reflect the sequence information between multiple subtitle sentences, we use BiGRU on \sh{both text and video} respectively. 
Then, we apply the max-pooling operation across the sequence of the subtitle sentences, to get a global representation of each answer, \sh{called hypothesis:} 
\begin{equation}
\begin{split}
\mathcal{H}_v^{Att} =& \;\;\text{Max}(\text{BiGRU}(H_v^{Att})), \\
\mathcal{H}_t =&  \;\;\text{Max}(\text{BiGRU}(H_t)). \\
\end{split}
\label{eq:bigru}
\end{equation}

Given the max-pooled hypothesis representations, we use two fully-connected layers as classifiers to obtain the logits \nj{$s_t$ and $s_v$} for the answer options on both sides \sh{of text and video} respectively. 

\begin{equation}
\begin{split}
s_v =\;\; \text{classifier}(\mathcal{H}_v^{Att}), \quad 
s_t =\;\; \text{classifier}(\mathcal{H}_t).
\end{split}
\label{eq:classifier}
\end{equation}

Then, we add those logits followed by a softmax function to obtain a probability distribution of each answer option and apply cross-entropy loss as our question answering loss:
\begin{equation}
\begin{split}
\hat{y} = \text{softmax}(s_v + s_t), \quad
\mathcal{L}_{qa} = - \sum_{i=1}^{N}  y_i \log{\hat{y}_i}.
\end{split}
\label{eq:loss_cls}
\end{equation}

\subsubsection{Temporal Localization}

We use temporal localization network, shown in Figure \ref{fig:arch}(b), which localizes relevant moments from a long video sequence given a question, and assign the ground truth start/end sentence position in the subtitle sequence using the given start/end time annotations.
We utilize the BiGRU output $\mathcal{H}_t$ from the text input, reflecting the sequence information of the text context and a question. Then, we predict the start/end position using span predicting classifiers \ksh{followed by a max-pooling operation across the five hypotheses}, and train them with cross-entropy loss as follows:
\begin{equation}
\begin{split}
\mathcal{L}_{span} = -\frac{1}{2}(\log{p_{start}}+\log{p_{end}})
\end{split}
\label{eq:loss_span}
\end{equation}
where $p_{start}$ and  $p_{end}$ are the span probabilities of the start and end ground truth positions respectively. Since we use this temporal localization part as one of our auxiliary tasks, we do not need start/end time annotations as well as temporal localization network in the inference time.

\begin{table}[t]
\resizebox{\linewidth}{!}
{
\small
\begin{tabular}{p{2.8in}}
	\hline
	\textbf{Conceptual text input} \\
	\vspace{0.1mm}
    {} $\mathtt{[CLS]}$ {\em question} $\mathtt{[SEP]}$ {\em answer option} $\mathtt{[SEP]}$ {\em subtitle sentence} $\mathtt{[SEP]}$ {\em objects} $\mathtt{[SEP]}$\\
	\hline
	\textbf{Original text input} \\
	\vspace{0.1mm}
    {} $\mathtt{[CLS]}$ {\em Where does Ted go after leaving the bar ?} $\mathtt{[SEP]}$ {\em Ted goes to Marshall's apartment 
        to tell him about the trip} $\mathtt{[SEP]}$ {\em Marshall : In fact, take my car .} $\mathtt{[SEP]}$ {\em necklace brown shirt woman ...} $\mathtt{[SEP]}$\\
	\hline
    \textbf{Masked text input} \\
	\vspace{0.1mm}
    {} $\mathtt{[CLS]}$ {\em Where does Ted go } $\mathtt{[MASK]}$ {\em leaving the bar ?} $\mathtt{[SEP]}$ {\em Ted } $\mathtt{[MASK]}$ {\em to Marshall's } $\mathtt{[MASK]}$  {\em  to tell him about the trip} $\mathtt{[SEP]}$ $\mathtt{[MASK]}$ {\em : In fact, take my car .} $\mathtt{[SEP]}$ {\em necklace brown } $\mathtt{[MASK]}$ {\em woman ...} $\mathtt{[SEP]}$\\
	\hline
	\textbf{Answer-removed text input} \\
	\vspace{0.1mm}
    {} $\mathtt{[CLS]}$ {\em Where does Ted go after leaving the bar ?} $\mathtt{[SEP]}$ $\mathtt{[MASK]}$ $\mathtt{[SEP]}$ {\em Marshall : In fact, take my car .} $\mathtt{[SEP]}$ {\em necklace brown shirt woman ...} $\mathtt{[SEP]}$\\
	\hline	
\end{tabular}
}
\caption{Examples of text input of BERT. Original text input is used in a QA network, masked text input is used in a contrastive learning network, and answer-removed text input is used in a self-supervised pre-training stage. Note that, for the readability, we do not use subword tokens \nj{in} these examples.}
\label{tab:masked_input}
\end{table}

\subsubsection{Contrastive Learning}

Figure \ref{fig:arch}(c) shows a proposed contrastive learning approach, as another auxiliary task, that enhances the model performance in the multiple-choice video QA.
\sh{As described by the masked text input example in Table \ref{tab:masked_input}}, we first mask out the tokens of the text input, \sh{corresponding to} the ground truth answer, with a certain probability using a special token $\mathtt{[MASK]}$. 
We encode the masked text input using the same BERT and BiGRU, that are used in the video QA section (Figure 2(a)), and denote the encoded representation as an anchor, $\mathcal{H}_{anchor}\in \mathbb{R}^{1 \times 2d_t}$.
Then, we employ contrastive learning, comparing masked anchor representation and previously extracted \nj{text} representations, $\mathcal{H}_t\in \mathbb{R}^{N \times 2d_t}$ in eq. (\ref{eq:bigru}) from the video QA network.
In the representations from the video QA network, we consider the representation 
corresponding to the ground truth answer as a positive sample, and others as negative samples, and use the dot product to measure the similarity scores \nj{between the text representation and} the anchor representation.
\begin{equation}
\begin{split}
scores = \mathcal{H}_t\mathcal{H}_{anchor}^T
\end{split}
\label{eq:cont_dot}
\end{equation}
Then, we apply the softmax \sh{to the computed} similarity scores and optimize it with the cross-entropy loss that can contrast the positive and negative representations correctly.

\begin{equation}
\begin{split}
\hat{y}_{con} = \text{softmax}(scores), \ 
\mathcal{L}_{cont} = - \sum_{i=1}^{N}  y_i \log{\hat{y}_{con, i}} 
\end{split}
\label{eq:loss_cont}
\end{equation}

Finally, \sh{introducing scale parameters; $\lambda_{qa}$, $\lambda_{span}$, and $\lambda_{cont}$}, the total loss is \nj{defined as} a linear combination of the above three losses as follows:
\begin{equation}
\begin{split}
\mathcal{L}=\lambda_{qa}*\mathcal{L}_{qa}+\lambda_{span}*\mathcal{L}_{span}+\lambda_{cont}*\mathcal{L}_{cont}
\end{split}
\label{eq:loss_total}
\end{equation}

\begin{table*}[t]
\centering
\resizebox{0.92\textwidth}{!}{
\begin{tabular}{l | c | c c c c c c c}
\hline\hline
\multirow{2}{*}{\textbf{Models}}  & \textbf{Val (Acc.)}  & \multicolumn{7}{c}{\textbf{Test-public (Acc.)}} \\

& All & bbt  & friends  & himym  & grey & house & castle & All  \\  \hline
multi-stream \cite{lei2018tvqa} & 65.85 & 70.25 & 65.78 & 64.02 & 67.20 & 66.84 & 63.96 & 66.46 \\
PAMN \cite{kim2019progressive} &   66.38 & -  & -  & -  & - & - & -  & 66.77  \\ 
Multi-task  \cite{kim2019gaining} & 66.22 & -  & -  &-   &-  & - & -  & 67.05 \\ 
CA-RN  \cite{geng2020character} & 68.90 & 71.43  & 65.78  & 67.20  & 70.62 & 69.10  & 69.14  & 68.77 \\ 
STAGE  \cite{lei2019tvqa+} & 70.50 & -  & -  & -  & - & - & -  & 70.23 \\ 
akalsdnr (anonymous) &  71.13 & 71.49  & 67.43  & 72.22  & 70.42 & 70.83 & 72.30  & 70.52 \\ 
MSAN  \cite{kim2020modality} &  70.79 & -  & -  & -  & - & - & -  & 71.13 \\ 
DenseCap  \cite{kim2020densecaption} & 74.20 &  74.04 & 73.03  &  74.34 & 73.44 & 74.68 &  74.86 & 74.09  \\ 

\hline
\textbf{Ours} & \textbf{76.23} & \textbf{77.43}  & \textbf{73.24} & \textbf{76.72} & \textbf{74.04} & \textbf{76.94} & \textbf{77.86} & \textbf{76.15}  \\
\hline \hline
\end{tabular}
}
\caption{Comparison of QA performance with previous methods on TVQA validation and test sets. All results are from the models that do not use timestamp annotations (w/o ts version). We also compare the performance on the 6 individual TV shows.}
\label{tab:perf_tvqa}
\end{table*}

\subsubsection{Self-supervised Pre-training}

We propose a self-supervised pre-training approach that is applicable to the multiple-choice video QA task.
While the original problem is to predict the answer using a question and text-visual contexts as eq. (\ref{eq:problem}), we \sh{instead train the model to} predict the corresponding question using text-visual contexts. 
\begin{equation} 
\hat{q} = \argmax_{q\in \Omega_q} p(q|S,V;\theta) 
\label{eq:reverse_problem}
\end{equation}
\sh{In this pre-training stage, we randomly sample negative questions for given video clip to learn the question-video alignment. For each negative training sample, since we previously know which video clips contain which questions, we select questions from other video clips that are \nj{not} related to the given video clip.}
In this process, we do not need correct answer annotation, since we replace the part corresponding to the answer \ksh{option} in the input to a single $\mathtt{[MASK]}$ token as follows:
\begin{quote}
\centering 
$\mathtt{[CLS]}$ $q_n$ $\mathtt{[SEP]}$ $\mathtt{[MASK]}$ $\mathtt{[SEP]}$ $S_t$ $\mathtt{[SEP]}$ $o_t$ $\mathtt{[SEP]}$
\end{quote}
\sh{The answer-removed text input example in Table \ref{tab:masked_input} shows an example used in the self-supervised pre-training stage.}
And as with the main stage, not only question answering loss but also temporal span and contrastive losses are also used in the pre-training stage as eq. (\ref{eq:loss_total}). 
By predicting which question comes from a given context, our proposed network can learn stronger representation with a better parameter initialization to improve the model performance.

\section{Experiments}
We evaluate our approach on three benchmark datasets: TVQA, TVQA+ and DramaQA.
TVQA is a large scale multiple-choice video QA dataset based on 6 popular TV shows: \textit{The Big Bang Theory, How I Met Your Mother, Friends, Grey's Anatomy, House, Castle}, and consists of 152,545 QA pairs from 21,793 clips, spanning over 460 hours of video. The training, validation, and test-public set consist of 122,039, 15,253, and 7,623 questions, respectively.
TVQA+ is a subset (\textit{The Big Bang Theory}) of TVQA, but TVQA+ adds frame-level bounding box annotations for visual concept words and modifies its timestamp information for better annotations. TVQA+ contains 29,383 QA pairs from 4,198 video clips, with 148,468 images annotated with 310,826 bounding boxes. The training, validation, and test-public set consist of 23,545, 3,017, and 2,821 questions, respectively. Note that we do not use bounding box information on TVQA+ to match the problem format \nj{to that} of TVQA.
DramaQA is built upon the TV drama (\textit{Another Miss Oh}) and it contains 16,191 QA pairs from 23,928 various length video clips. The QA pairs \nj{belong} to one of four difficulty levels and \nj{the dataset} provides the character-centered annotations, including visual bounding boxes, behaviors, and emotions of main characters. 
\ksh{\nj{As in TVQA+}, we do not use bounding box information at all. However, we use textual information regarding behaviors and emotions as objects.}
\nj{The} number of examples for training, validation, and test datasets \nj{is} 10,098, 3,071, and 3,022, \nj{respectively}.

\subsection{Implementation Details}

We use pre-extracted \nj{2048-\sh{dimensional}} hidden features ($d_v$ in Fig. \ref{fig:arch}) from \nj{the Imagenet-pretrained ResNet-101} and object information from the modified Faster R-CNN trained on Visual Genome \cite{lei2018tvqa}.
We use \nj{the} BERT-base uncased model, which has 12 layers with hidden size \sh{of 768} and fine-tuned only top-6 layers due to the limitation of resources. 
We set the hidden sizes of all the remaining layers as 768 ($d_t$ in Fig. \ref{fig:arch}).
The total video context sequence $T$ is 40, the number of images $I$ corresponding to each subtitle sentence is \sh{set to} 4, and the number of answer options $N$ is 5 in \ksh{all} datasets, as shown in Figure \ref{fig:arch}. The maximum number of tokens of the text input is set to 80 for TVQA/TVQA+ and 170 for DramaQA. 
The probability of masking out the tokens \sh{used in our contrastive learning} is 0.2.
The weights of each loss $\lambda_{qa}$, $\lambda_{span}$, and $\lambda_{cont}$ are set to 1, 0.2, and 0.1 based on TVQA+ validation performance. We set the learning rate to 1e-5 for the self-supervised pre-training stage and 5e-5 for the main QA stage. Likewise, the total number of epochs is set to 1 for the pre-training stage and 3 for the main QA stage. \sh{We use the batch size of 8 for the entire experiment settings.}

\subsection{Experimental Results}

\subsubsection{TVQA}

We evaluate our model on TVQA dataset as shown in Table \ref{tab:perf_tvqa}.
Since the ground truth answers of the test set are not provided, we \sh{present the performance via the online evaluation server system.}
Our model achieves 76.15\% of accuracy on the test set, outperforming the previous state-of-the-art models,
MSAN \cite{kim2020modality} using modality importance with BERT and DenseCap \cite{kim2020densecaption} using captions and frame-selection with RoBERTa, with over 5\% and 2\% margins, respectively. Our model also achieves the best performance in all 6 individual TV shows.

\subsubsection{TVQA+}

Table \ref{tab:perf_tvqa+} shows the performance on TVQA+ dataset. To measure \sh{the performance of} our model, we use QA classification accuracy \sh{just like in} TVQA, and additionally, temporal mean Intersection-over-Union (mIOU) and Answer-Span joint Accuracy (ASA) provided by \citet{lei2019tvqa+} are used.
mIOU measures temporal localization and ASA jointly evaluates \sh{the performance of} both QA classification and temporal localization. For the ASA metric, we regard a prediction to be correct if the predicted temporal localized span has an IoU $\geq$ 0.5 with the ground-truth span and the answer is correctly predicted.
We \nj{obtained} the accuracy of 76.21\% \sh{in QA classification} and the mIoU of 39.03\% in temporal localization.
\sh{For ASA,} 
we achieved 31.05\%, outperforming the previous state-of-the-art model, STAGE \cite{lei2019tvqa+} which used BERT with grounding spatial regions and temporal moments, with about 9\% margin.

\subsubsection{DramaQA}

Table \ref{tab:perf_dramaqa} shows our result on DramaQA dataset, consisting of four levels of difficulty.
\sh{The first five lines} of Table \ref{tab:perf_dramaqa} show the top-5 resulting models of the DramaQA challenge, evaluated on the test set. \sh{Since the challenge is no longer ongoing and the test set is yet inaccessible,} we evaluate our model only on the \sh{available} validation set and report ours for future benchmark comparison. Although direct comparison is difficult, our model \sh{shows competitive performances among others.}

\begin{table}[t]
\centering
\resizebox{\linewidth}{!}
{
\begin{tabular}{lccc}
	\hline\hline
	\textbf{Models} & \textbf{QA (Acc.)}  & \textbf{TempLocal (mIOU)}  & \textbf{ASA} \\
	\hline
    ST-VQA \cite{jang2017tgif} &	 48.2 &	- & - \\
    two-stream \cite{lei2018tvqa} &	 68.13 & -&-  \\
    STAGE (video) \cite{lei2019tvqa+} &	52.75 &	 10.90 & 2.76 \\
    STAGE (sub) \cite{lei2019tvqa+} &	67.99 &	 30.16 & 20.13 \\
    STAGE \cite{lei2019tvqa+} &	74.83 &	 32.49 & 22.23 \\
  	\hline
	\textbf{Ours} & \textbf{76.21} & \textbf{39.03} & \textbf{31.05} \\
	\hline\hline
\end{tabular}
}
\caption{Comparison \nj{on} TVQA+ test set. We evaluate QA accuracy, mIoU for temporal localization, and Answer-Span joint Accuracy (ASA) as the overall performance \ksh{indicator\nj{s}}.}
\label{tab:perf_tvqa+}
\end{table}

\begin{table}[t]
\centering
\resizebox{\linewidth}{!}
{
\begin{tabular}{lccccc}
	\hline\hline
	\textbf{Models} & \textbf{Difficulty 1}  & \textbf{Difficulty 2}  & \textbf{Difficulty 3}  & \textbf{Difficulty 4}  & \textbf{Overall} \\
	\hline
    IITDrama &	 76 &	72 & 55 & 60 & 71 \\
    bjorn  &	 77 &	74 & 57 & 57 & 71 \\
    HARD KAERI  & 76&	73 & 56 & 59 & 71 \\
    Sudoku &	 78 &	74 & 68 & 67 & 75 \\
    GGANG&	 81 &	79 & 64 & 70 & 77 \\
  	\hline
	\textbf{Ours (validation)} & 84 & 85  & 70 & 70 & 81 \\
	\hline\hline
\end{tabular}
}
\caption{QA accuracy on DramaQA dataset with four difficulty levels. \sh{Task becomes more difficult as the level increases.} 
We report top-5 results from the competition leaderboard, evaluated on the test set. Note that, we only evaluate on the validation set since the challenge is no longer ongoing and the test set is yet inaccessible.}
\label{tab:perf_dramaqa}
\end{table}

\begin{table}[t]
\centering
\resizebox{0.7\linewidth}{!}
{
\small
\begin{tabular}{lc}
	\hline\hline
	\textbf{Models} & \textbf{QA (Acc.)} \\
	\hline
    (1) base model (GA) & 71.62 $\pm$ 0.45 \\
    (2)  \hspace{2mm} + TL &	 73.45 $\pm$ 0.31  \\
    (3)  \hspace{2mm} + TL + MT &	 73.98 $\pm$ 0.27  \\
	\hline
    (4) base model (LA) & 72.29 $\pm$ 0.31 \\
    (5)  \hspace{2mm} + TL &	73.53 $\pm$ 0.29 \\
    (6)  \hspace{2mm} + TL + MT &	 74.54 $\pm$ 0.21  \\
	\hline
    (7)  \hspace{2mm} + TL + MT + CL &	 75.16 $\pm$ 0.18 \\
    (8)  \hspace{2mm} + TL + MT + CL + SS &	 75.83 $\pm$ 0.06  \\
	\hline\hline
\end{tabular}
}
\caption{Results of the ablation study of our model \nj{on TVQA+ validation set}.
We ablate our model with globally aligned attention (GA), locally aligned attention (LA), multiple token type embeddings (MT), Temporal localization span loss (TL), contrastive loss (CL), and self-supervised pre-training stage (SS).}
\label{tab:ablation}
\end{table}

\subsection{Analysis}

\subsubsection{Ablation study}
We conduct an ablation study on the TVQA+ \sh{validation} set as shown in Table \ref{tab:ablation}.
For an ablation experiment, we define base models where \sh{the token type embedding, temporal localization, contrastive learning, and self-supervised stage are removed.}
The base models consist of the globally aligned attention model (1) and the proposed locally aligned attention model (4). 
First, we observe that the models with locally aligned attention outperformed all the \sh{other} models \sh{that are trained with a} globally aligned attention in \nj{(1-3 vs. 4-6)}. It \sh{implies} that \sh{misalignment between subtitle sentences and the image frames from other sentences can be prevented} by utilizing the time sequence information.
Second, \sh{(2-3,5-6) show the effectiveness of the proposed multiple token type embedding technique,} and we believe this \sh{can be extensively applied when working with} various types of text inputs.
In (7), we use the proposed contrastive learning with the masked text input as the auxiliary task. This brings additional performance improvement from 74.54\% to 75.16\% of accuracy.
Lastly, using self-supervised pre-training with a \ksh{transformed} problem format as a prerequisite learning, we achieve the best performance of 75.83\% accuracy on TVQA+ validation set. 
It demonstrates that \sh{our model takes further advantage of the given dataset using the self-supervised pre-training scheme.}

\begin{figure}[t]
  \includegraphics[width=\linewidth]{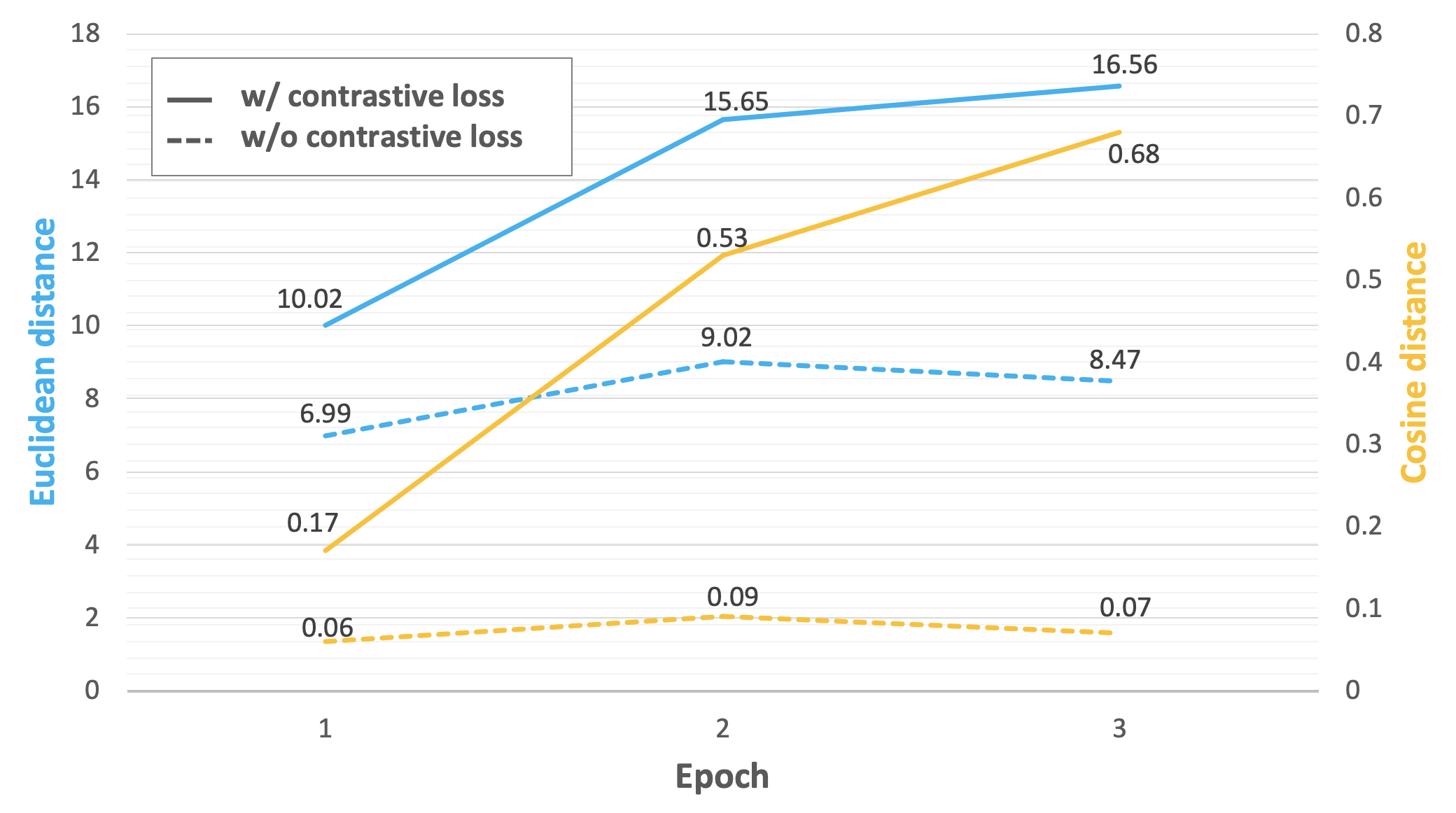}
  \caption{Euclidean and Cosine distances between the positive representation and the closest negative representation from the positive one according to whether or not the contrastive loss is used.}
\label{fig:distance}
\end{figure}

\subsubsection{Effectiveness of the \sh{proposed} contrastive loss}

For the contrastive representation learning, as shown in Fig. \ref{fig:arch}(c), \nj{among five QA pairs, we contrast a single ground truth answer with the other four  negative answers.}
We investigate how the hidden representations ($\mathcal{H}_t$ in eq. (\ref{eq:bigru})) of the five QA pairs (one positive and four negatives) \sh{behave} depending on \sh{whether the contrastive loss is used or not.}
We calculate the distance between the positive and the closest negative representation\sh{s} and \sh{report} how the distance \sh{between them is changing} as the epoch continues.
For the distance metric, we use the euclidean and cosine distance functions. Figure \ref{fig:distance} shows that \sh{the distance between the positive and the closest negative representations is increasing in both metrics when the contrastive loss is accompanied during the training,} while there is no noticeable increase in distance when the contrastive loss is not used. This shows that applying the proposed contrastive loss helps to separate the representation space between the positive and negative \sh{samples} and we believe \ksh{these separated} representations are helpful for predicting the answer correctly.

\subsubsection{Qualitative Results}

\begin{figure}[t]
    \centering
    \includegraphics[width=1.\linewidth]{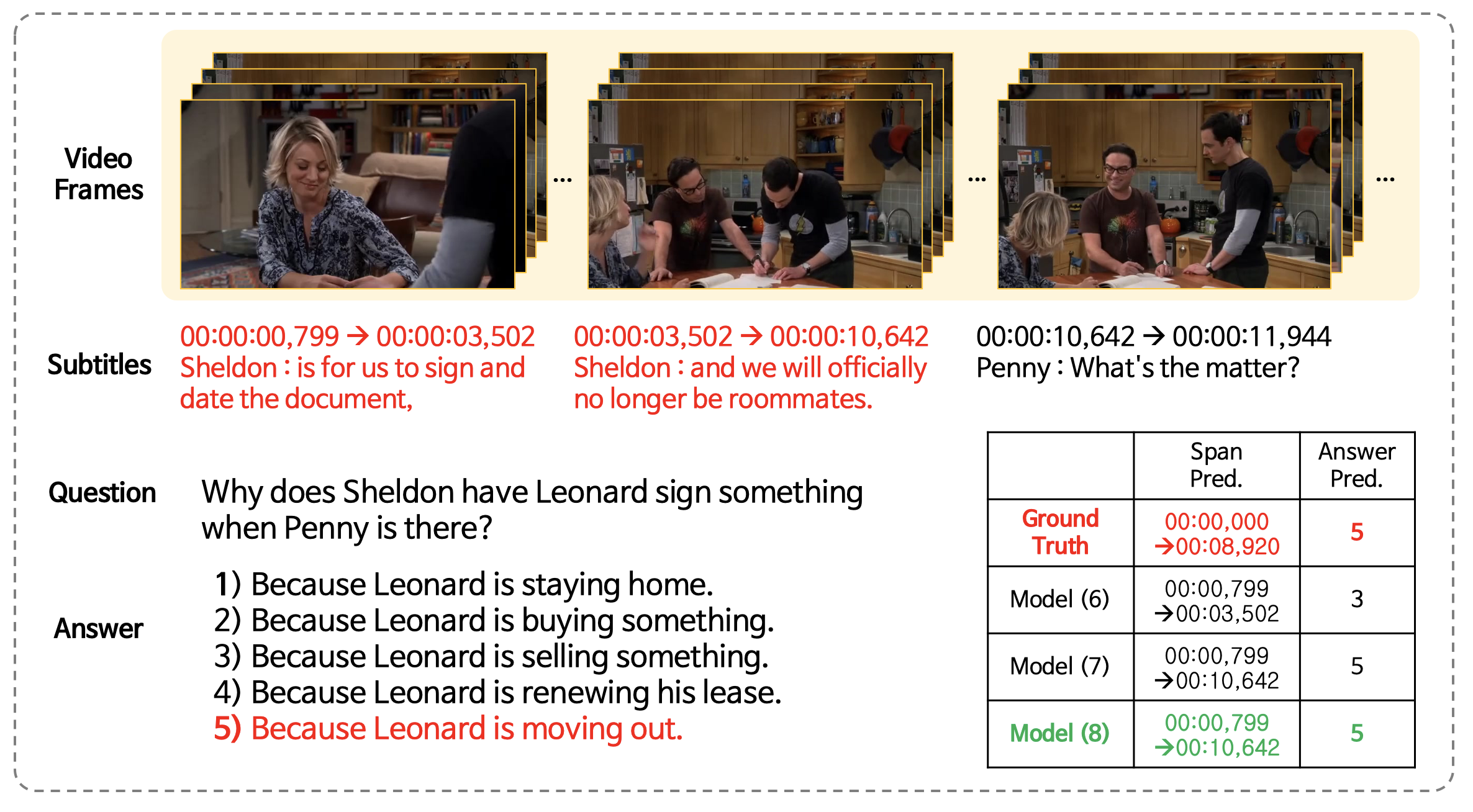}
    \\
    \includegraphics[width=1.\linewidth]{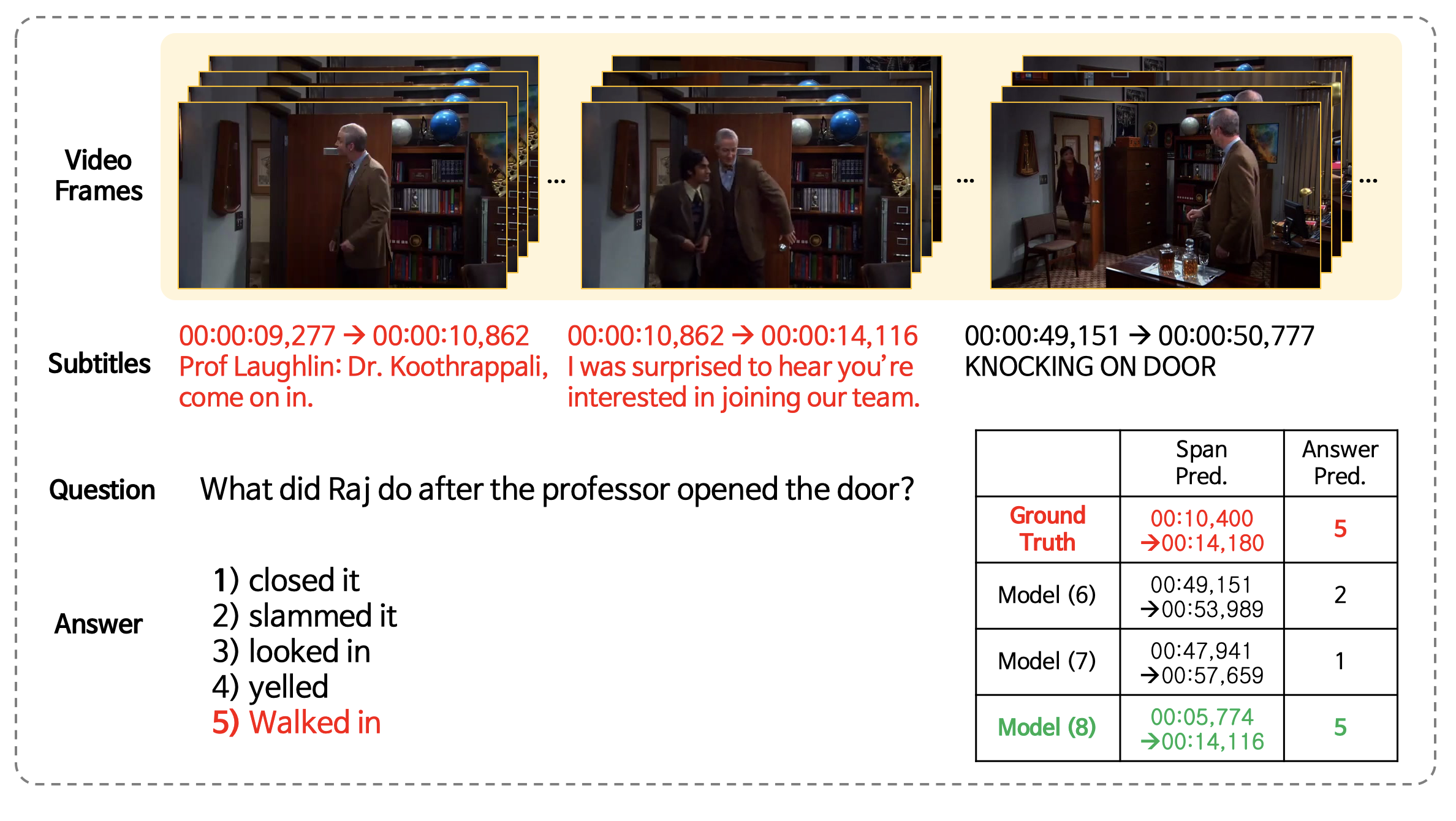}
    \caption{Examples of predictions of models with or without the contrastive loss and the self-supervised pre-training scheme.
    The ground truths are denoted in red, and the predictions of our proposed model are colored in green.}
    \label{fig:sample}
\end{figure}

Figure \ref{fig:sample} shows two examples of the prediction according to the use of the proposed contrastive representation learning and the self-supervised pre-training (model 6-8 in Table \ref{tab:ablation}). 
In the first example, the proposed model predicts the correct answer by \sh{associating Sheldon's dialog}, ``officially no longer be roommates", with the expression ``moving out" in the correct answer and gives 0.76 of IoU in the temporal localization, while the model without two proposed approaches (model (6)) predicts the wrong answer with only 0.3 of IoU with the ground truth video span.
The second example requires both language and visual understanding to predict the answer and the video span correctly.
Our final model localizes the related video span and predicts the answer correctly.
However, other models rather pay attention to the word ``door" which appear in both of the question and the subtitle sentence and fail to predict the correct answer.

\section{Conclusion}
\sh{Video QA requires fine-grained understanding of both video and language modalities. To address this, we focus on the training procedure that could possibly take the most advantage out of the given data.}
In this paper, we propose novel training schemes \sh{that} specialize in multiple-choice video QA. We first pre-train our model with a transformed problem format \sh{of predicting which questions are from which contexts} for a better weight initialization.
\sh{We} then train \sh{our} model with the original QA problem format while being guided by contrastive representation learning. Our model achieves state-of-the-art performance on three highly challenging video QA datasets.
We expect that our proposed method can be applied for various multiple-choice video QA tasks, bringing further performance improvement.

\section*{Acknowledgments}
This work was supported by IITP grant funded by the Korea government (MSIT) (No.2019-0-01367, Babymind) and Next-Generation Information Computing Development Program through the NRF of Korea (2017M3C4A7077582).

\bibliography{vqa21}
\end{document}